\documentclass[a4paper,conference]{IEEEtran}

\usepackage{cite}
\usepackage{soul} 
\usepackage{color} 
\usepackage{amsmath,amssymb,amsfonts}
\usepackage{algorithmic}
\usepackage{graphicx}
\usepackage{textcomp}
\usepackage{xcolor}
\usepackage{url}
\usepackage{multirow}
\usepackage{float} 
\usepackage{pgfplots} 
\usetikzlibrary{pgfplots.statistics}
\pgfplotsset{every tick label/.append style={font=\footnotesize}}
\def\BibTeX{{\rm B\kern-.05em{\sc i\kern-.025em b}\kern-.08em
    T\kern-.1667em\lower.7ex\hbox{E}\kern-.125emX}}

\usepackage{fancyhdr}
\usepackage{eso-pic}

\begin{document}

\AddToShipoutPictureBG*{%
  \AtPageUpperLeft{%
    \setlength\unitlength{1in}%
    \hspace*{\dimexpr0.5\paperwidth\relax}
    \makebox(0,-0.75)[c]{\small Convention of the \emph{Society for the Study of Artificial Intelligence and Simulation of Behaviour} (AISB), 13-14 April 2023, Swansea, UK}%
    }}
    
\title{Local Minima Drive Communications in Cooperative Interaction}

\author{\IEEEauthorblockN{Roger K.\ Moore}
\IEEEauthorblockA{\textit{Dept.\ Computer Science} \\
\textit{University of Sheffield}\\
Sheffield, UK \\
r.k.moore@sheffield.ac.uk}
}

\maketitle

\begin{abstract}
An important open question in human-robot interaction (HRI) is precisely \emph{when} an agent should decide to communicate, particularly in a cooperative task.  Perceptual Control Theory (PCT) tells us that agents are able to cooperate on a joint task simply by sharing the same `intention', thereby distributing the effort required to complete the task among the agents.  This is even true for agents that do not possess the same abilities, so long as the goal is observable, the combined actions are sufficient to complete the task, and there is no local minimum in the search space.  If these conditions hold, then a cooperative task can be accomplished \emph{without} any communication between the contributing agents.  However, for tasks that \emph{do} contain local minima, the global solution can only be reached if at least one of the agents adapts its intention at the appropriate moments, and this can only be achieved by appropriately timed communication.  In other words, it is hypothesised that in cooperative tasks, the function of communication is to coordinate actions in a complex search space that contains local minima.  These principles have been verified in a computer-based simulation environment in which two independent one-dimensional agents are obliged to cooperate in order to solve a two-dimensional path-finding task.
\end{abstract}


\section{Introduction}
An important open question in human-robot interaction (HRI) is precisely \emph{when} an agent should decide to communicate \cite{Skantze}.  Unfortunately, research in human-human interaction has been obsessed with `turn-taking' as the underlying mechanism \cite{Sacks,Wilson,Gravano,Levinson}, somewhat overlooking the observation that conversation can overlap as well as interleave \cite{Schegloff,Heldner}, as well as ignoring the question as to \emph{why} agents should communicate in the first place \cite{Cowley}.  Clearly, communication supports information exchange \cite{Goldman,Buehler} and learning \cite{Das}, but more importantly it facilitates collaborative problem solving \cite{Lazaridou} and goal sharing \cite{Xuan}, i.e.\ \emph{cooperation}.  However, little research has been conducted into what conditions the timing and structure of communication in continuous cooperative interaction \cite{Vesper}.

This paper addresses these issues from the perspective of Perceptual Control Theory (PCT) \cite{Powers}.  Results are presented from a PCT-based simulation of a cooperative task, and it is shown how appropriately timed communication between agents can overcome local minima in a joint problem space.

\section{Communication in Cooperation}

Perceptual Control Theory (PCT) is founded on the mantra ``\emph{behaviour is the control of perception}'', and agents are modelled as a hierarchy of negative-feedback control loops.  Solidly grounded in the tradition of `cybernetics' \cite{Wiener}, PCT has been shown to be capable of accounting for a wide range of `intelligent' phenomena based on a parsimonious architecture of replicated closed-loop structures \cite{Mansell}.  In particular, PCT tells us that agents are able to cooperate on a joint task simply by sharing the same reference signal, i.e.\ by having the same \emph{intention} \cite{Farrell}.  The consequence is that the effort required to complete a task may be distributed among the agents involved.

However, it is claimed here that successful convergence towards a solution of a joint task is based on three assumptions:
\begin{itemize}
	\item the goal is observable (that is, each agent has an appropriate input function),
	\item the combined actions are sufficient to complete the task (that is, the agents possess complimentary output functions), and
	\item the goal is accessible (that is, there are no \emph{local minima} in the search space).
\end{itemize}

If these three conditions are met, then a cooperative task can be accomplished \emph{without} any communication between the contributing agents.

This means that, for tasks that \emph{do} have local minima, the global solution can only be reached if at least one of the agents adapts its intention at the appropriate moment(s).  That is, an agent may need to abandon its original goal in favour of a temporary alternative that facilitates an escape from a local minimum.  Such behaviour requires timely coordination between the agents, and this can only be achieved by appropriately-timed \emph{communication}.  In other words, it is hypothesised that, in cooperative tasks, one function of communication is to coordinate actions in a complex search space that contains one or more local minima.  From a PCT perspective, this implies that a perceived signal from one agent should trigger a change in a reference signal for another agent.

This hypothesis has been verified in a computer-based simulation in which two independent one-dimensional agents are obliged to communicate (that is, actively cooperate) in order to solve a two-dimensional path-finding task \cite{Moore2022}.

\section{Simulation Environment}

\begin{figure}[h!]
	\centering
	\includegraphics[width=\columnwidth]{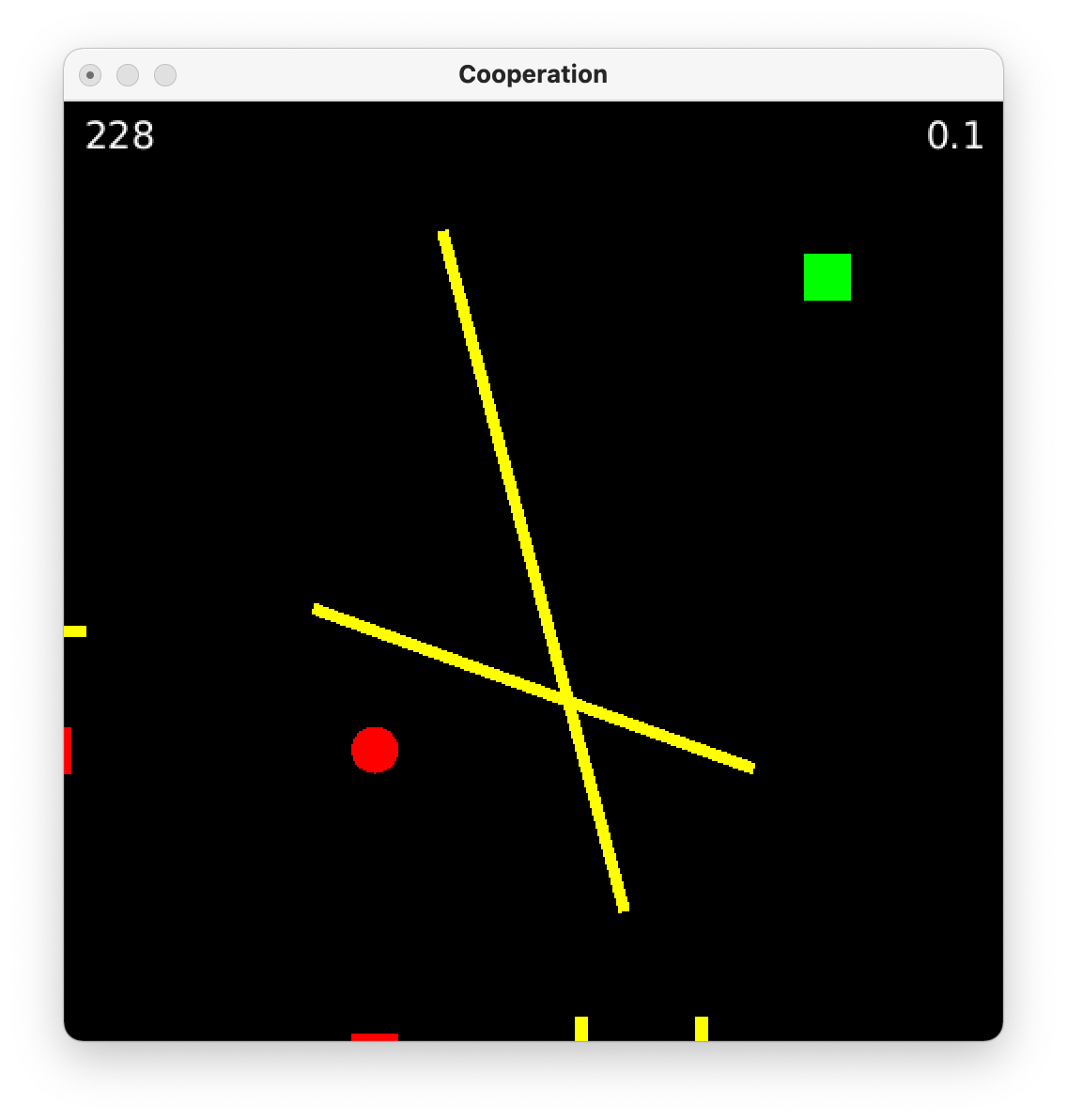}
	\caption{Screenshot of the Pd-based simulation environment showing the target (green square), the vehicle (red circle) and two barriers (yellow lines).  The X and Y axes depict the 1D projection of the vehicle, target and barriers (if visible from the agent's perspective).  The number in the top-right corner indicates the elapsed time in the current run, and the number in the top-left corner shows the number of runs completed.}
	\label{fig:SimEnv}
\end{figure}

The simulation environment -- implemented in the Pure Data (Pd) dataflow programming language \cite{Pd,Moore2021} -- is illustrated in \figurename~\ref{fig:SimEnv}.  Two 1D agents control the X and Y positions of a `vehicle' in a 2D space, the task being to steer the vehicle towards a `target' location.  Each agent can only `see' the target in its single dimension, hence cooperation \emph{may} be required to solve the joint 2D problem.   The difficulty of the task is scaled by the introduction of various forms of obstruction (as illustrated in \figurename~\ref{fig:SimEnv}), and the `solution time' (ST) for each successful run was measured.

\begin{figure}[h!]
	\centering
	\includegraphics[width=0.9\columnwidth]{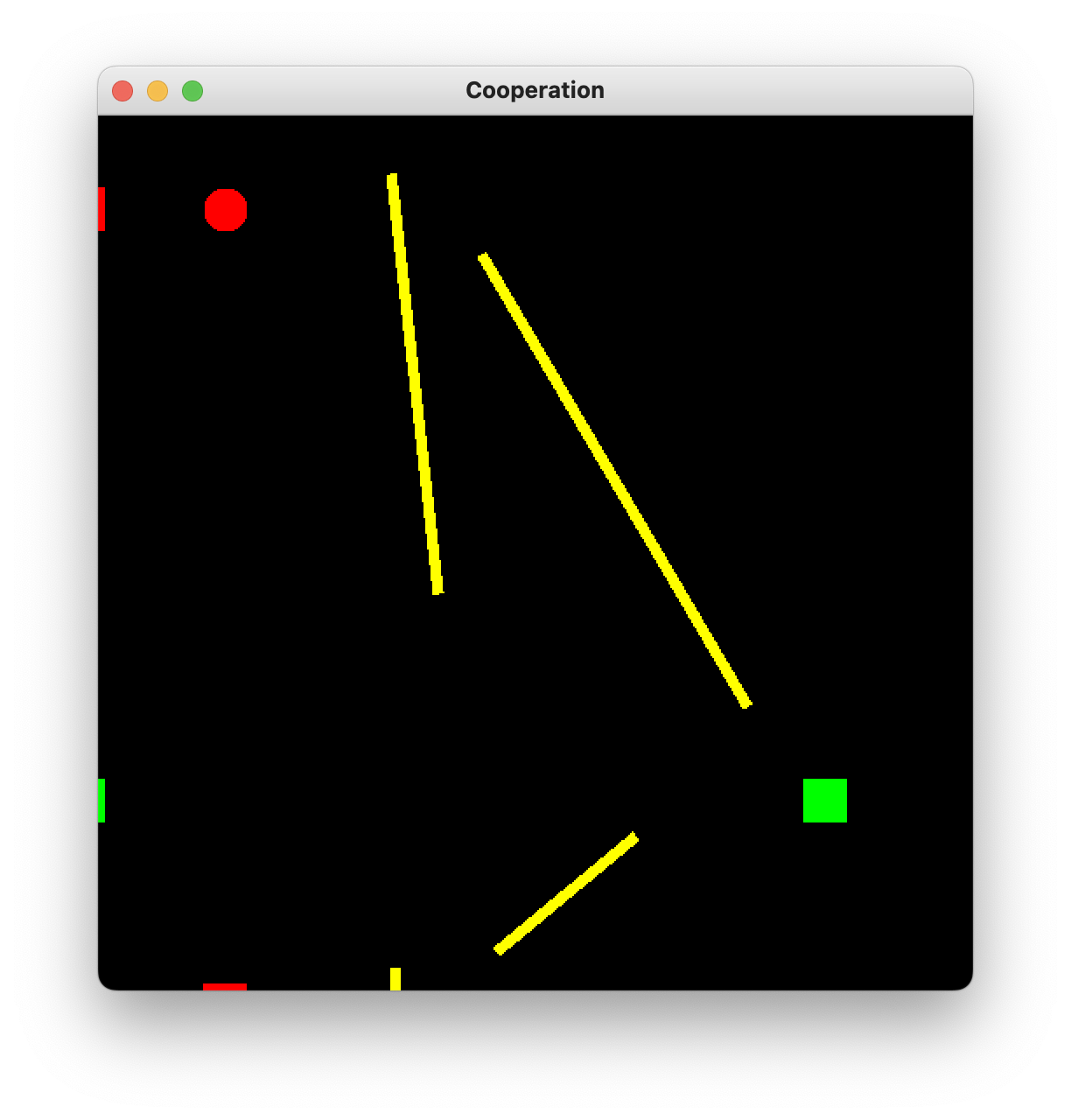}
	\caption{Screenshot of the Pd-based simulation environment showing a configuration with three barriers but \emph{no} local minimum.  This means that the vehicle can reach the target without getting `stuck'.}
	\label{fig:NoLM}
\end{figure}

\begin{figure}[h!]
	\centering
	\includegraphics[width=0.9\columnwidth]{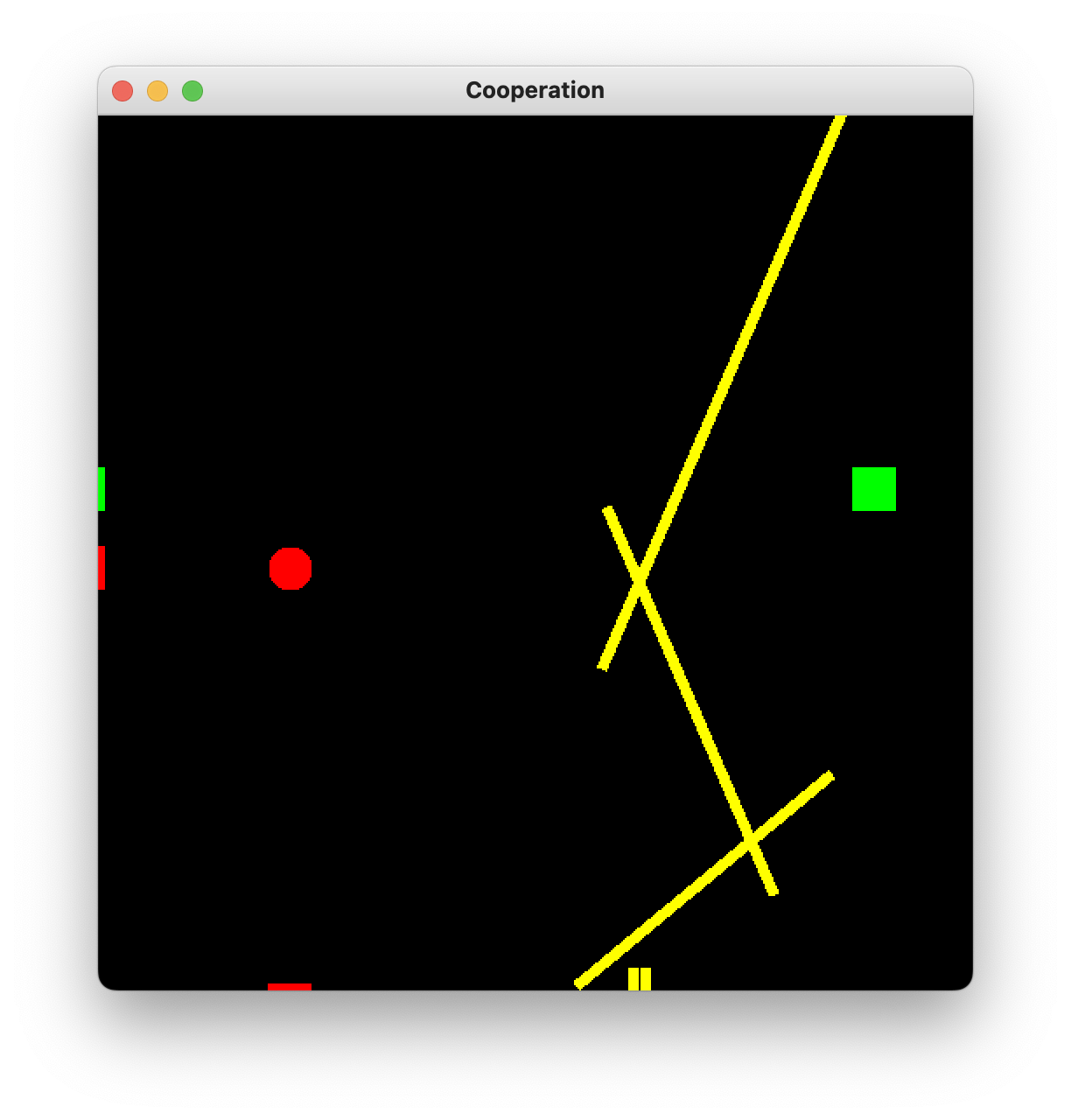}
	\caption{Screenshot of the Pd-based simulation environment showing a configuration with three barriers in which it is impossible for the vehicle to reach the target.}
	\label{fig:NoSol}
\end{figure}

There are many configurations in which each controlling X and Y agent can move the vehicle towards the target by reducing their individual `error' in a monotonic fashion (that is, by gradient descent), even if there are barriers present.  For example, \figurename~\ref{fig:NoLM} shows a configuration with three barriers but \emph{no} local minimum.  Also, some barrier configurations create situations in which it is impossible for the vehicle to reach the target at all -- see \figurename~\ref{fig:NoSol}.

However, some configurations (such as the one shown in \figurename~\ref{fig:SimEnv}) create situations which require agents to \emph{increase} their error momentarily in order for the vehicle to eventually reach the target.  For example, if one agent has reached its target (in 1D), but the other is stuck behind a barrier, then the first needs to be requested to abandon its target temporarily in an attempt to free the second agent.  Hence, the presence/absence of timely communications is critical in determining whether a run is ultimately successful or not.

Since not all configurations are solvable, the simulation environment was set up such that any experimental run lasting longer than 30 seconds was terminated and marked as `did not finish' (DNF).  In such cases, the solution time was ignored in subsequent data analysis.

Table~\ref{tab:setup} lists the variables instantiated in the simulations.  Overall, five cooperation modes were implemented \emph{per agent} (listed at the bottom of Table~\ref{tab:setup}), and different combinations were able to be specified by means of an agent-specific 4-bit binary code.  This meant that there was a total of 16 possible levels of cooperation available for each agent.  Two of these involved no communication at all, but distinguished between just stopping at an obstruction (i.e.\ no active cooperation) versus moving randomly (i.e.\ potential cooperation \emph{without} communicating).

\begin{table}[h!]
	\caption{Simulation Environment Setup}
	\centering
	\begin{tabular}{|c|l|}
		\hline
		\multicolumn{2}{|c|}{\textbf{Agent Variables}} \\
		\hline
		\multirow{1}{*}{\textbf{Action}}
		 & Forward, Reverse, Stop \\
		\hline
		\multirow{5}{*}{\textbf{Status}}
		& Collided with edge \\
		& Collided with barrier* (``\emph{stuck}'') \\
		& Target known \\
		& Target accessible* (``\emph{access}'') \\
		& Arrived at target (``\emph{arrived}'')  \\
		\hline \hline
		\multicolumn{2}{|c|}{\textbf{Experiment Variables}} \\
		\hline
		\multirow{5}{*}{\textbf{Environment}}
		& Initial target position [XY] \\
		& Initial vehicle position [XY] \\
		& Number of barriers [0--3] \\
		& Location, orientation \& size of barrier(s) [XYRL] \\
		& Number of runs [100--1000] \\
		\hline
		\multirow{4}{*}{\textbf{Agent}}
		& Target view [true/false] \\
		& Cooperation level [0000--1111]$^{\mathrm{a-e}}$ \\
		& Loop gain [0.01] \\
		& Back-off time [1000 msec] \\
		\hline
		\multirow{3}{*}{\textbf{Measures}}
		& Time to target (ST) [secs.] \\
		& Did not finish (DNF) [\#] \\
		& Time spent communicating [\%] \\
		\hline
		\multicolumn{2}{l}{$^{\mathrm{a}}$[0000] if self = target known $\Rightarrow$ approach target; else stop} \\
		\multicolumn{2}{l}{$^{\mathrm{b}}$[1000] if self = target known $\Rightarrow$ approach target; else move randomly} \\
		\multicolumn{2}{l}{$^{\mathrm{c}}$[0100] if self = ``\emph{arrived}'' + other = ``\emph{stuck}'' $\Rightarrow$ back-off} \\
		\multicolumn{2}{l}{$^{\mathrm{d}}$[0010] if self = ``\emph{stuck}'' + other = ``\emph{stuck}'' $\Rightarrow$ back-off} \\
		\multicolumn{2}{l}{$^{\mathrm{e}}$[0001] if self = ``\emph{access}'' + other = ``\emph{access}'' $\Rightarrow$ approach target}
	\end{tabular}
	\label{tab:setup}
\end{table}

Of particular interest are each agent's `status' parameters that were available to be communicated \emph{for a given level of cooperation}.  These are marked with a * in Table~\ref{tab:setup}.  

The first parameter -- ``\emph{stuck}'' -- relates to the identification of a potential local minimum.  Such a condition arises when one agent has collided with a barrier and the other has arrived at the target, or when both agents have collided with barriers.  Crucially, it was realised that just one agent being stuck at a barrier is not sufficient evidence for a local minimum, as the other agent may be making progress which could resolve the problem.  

The second parameter -- ``\emph{access}'' -- relates to whether the target was accessible, i.e. there was no barrier between the agent and the target.  However, it is important to appreciate that such a condition does not guarantee a successful approach, as the target may subsequently become inaccessible for one agent due to the activities of the other agent.

\section{Experiments \& Results}

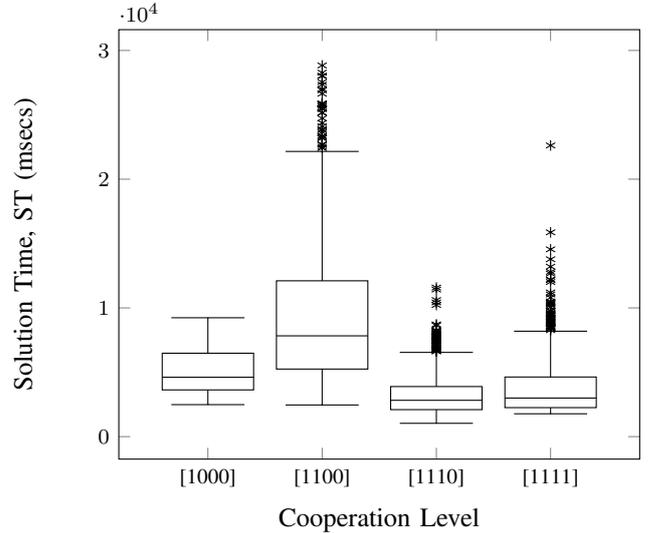
\begin{figure}[h!]
	\centering
	\begin{tikzpicture}
		\begin{axis}[boxplot/draw direction=y,xlabel=Cooperation Level,xtick={1,2,3,4},xticklabels={[1000],[1100],[1110],[1111]}, ylabel={Solution Time, ST} (msecs)]
    			\addplot+[boxplot,black,style={solid},mark options={fill=black},mark=asterisk]
			table[row sep=\\,y index=0] {
			data\\
3635 \\ 6270 \\ 4615 \\ 3650 \\ 4690 \\ 3830 \\ 5085 \\ 3245 \\ 3435 \\ 3625 \\ 6720 \\ 3225 \\ 5515 \\ 4010 \\ 7725 \\ 8815 \\ 4240 \\ 8025 \\ 3950 \\ 3295 \\ 5015 \\ 8685 \\ 9235 \\ 3005 \\ 7460 \\ 6405 \\ 6555 \\ 3565 \\ 4590 \\ 6710 \\ 4110 \\ 3155 \\ 2485 \\ 5005 \\ 7485 \\ 4735 \\ 3685 \\ 4975 \\
    };
			\addplot+[boxplot,black,style={solid},mark options={fill=black},mark=asterisk]
			table[row sep=\\,y index=0] {
			data\\
4100 \\ 8515 \\ 8265 \\ 7080 \\ 6120 \\ 3610 \\ 7385 \\ 8235 \\ 7515 \\ 5500 \\ 7190 \\ 5415 \\ 8425 \\ 3865 \\ 8785 \\ 7080 \\ 7800 \\ 17975 \\ 23200 \\ 2760 \\ 3840 \\ 4060 \\ 12095 \\ 13235 \\ 5945 \\ 5255 \\ 5805 \\ 8145 \\ 9515 \\ 6515 \\ 20100 \\ 9320 \\ 4875 \\ 4775 \\ 3425 \\ 11230 \\ 7545 \\ 23890 \\ 2835 \\ 8765 \\ 3945 \\ 7040 \\ 4875 \\ 16010 \\ 25150 \\ 15235 \\ 6280 \\ 4690 \\ 21625 \\ 20190 \\ 2720 \\ 17785 \\ 4740 \\ 4200 \\ 16860 \\ 4865 \\ 8915 \\ 4160 \\ 5750 \\ 5730 \\ 14210 \\ 28835 \\ 21635 \\ 12270 \\ 5305 \\ 7110 \\ 8285 \\ 10600 \\ 16055 \\ 19620 \\ 5840 \\ 16100 \\ 21050 \\ 11145 \\ 8795 \\ 10930 \\ 12270 \\ 5220 \\ 24110 \\ 28000 \\ 6275 \\ 4110 \\ 6160 \\ 15705 \\ 26905 \\ 10060 \\ 23655 \\ 8755 \\ 21370 \\ 13140 \\ 10460 \\ 3740 \\ 12400 \\ 5890 \\ 7695 \\ 6110 \\ 4140 \\ 9935 \\ 6395 \\ 11760 \\ 15910 \\ 4355 \\ 2790 \\ 16905 \\ 7395 \\ 4265 \\ 2715 \\ 3175 \\ 12945 \\ 4905 \\ 13855 \\ 8730 \\ 25175 \\ 9655 \\ 11385 \\ 4725 \\ 11895 \\ 3855 \\ 5725 \\ 8475 \\ 5740 \\ 4115 \\ 4735 \\ 10465 \\ 2755 \\ 28045 \\ 6530 \\ 8550 \\ 5170 \\ 7085 \\ 6040 \\ 25665 \\ 8715 \\ 4965 \\ 5950 \\ 4140 \\ 4285 \\ 4005 \\ 8265 \\ 11720 \\ 7225 \\ 23180 \\ 16415 \\ 16190 \\ 5440 \\ 4160 \\ 4655 \\ 6750 \\ 14565 \\ 9025 \\ 10880 \\ 8915 \\ 9230 \\ 14145 \\ 4115 \\ 10705 \\ 13165 \\ 3705 \\ 9370 \\ 13120 \\ 6505 \\ 3190 \\ 6825 \\ 4470 \\ 7485 \\ 13885 \\ 14815 \\ 9325 \\ 12950 \\ 3700 \\ 5605 \\ 7595 \\ 7215 \\ 6250 \\ 6965 \\ 3785 \\ 5560 \\ 3880 \\ 14260 \\ 5710 \\ 19160 \\ 15705 \\ 6640 \\ 5930 \\ 6195 \\ 11980 \\ 8315 \\ 4400 \\ 12765 \\ 7830 \\ 3720 \\ 9340 \\ 7640 \\ 5250 \\ 19985 \\ 4990 \\ 6195 \\ 18765 \\ 7850 \\ 18055 \\ 11530 \\ 9750 \\ 20675 \\ 21900 \\ 3640 \\ 13310 \\ 4450 \\ 21770 \\ 25845 \\ 12150 \\ 9310 \\ 15890 \\ 19790 \\ 4805 \\ 3825 \\ 12770 \\ 11815 \\ 3410 \\ 5705 \\ 5575 \\ 9310 \\ 18185 \\ 14465 \\ 4400 \\ 13060 \\ 14910 \\ 13390 \\ 4410 \\ 8260 \\ 5170 \\ 6820 \\ 16135 \\ 5580 \\ 8255 \\ 5745 \\ 5425 \\ 8415 \\ 8190 \\ 5210 \\ 4215 \\ 6430 \\ 10045 \\ 7000 \\ 5215 \\ 12350 \\ 6840 \\ 4510 \\ 9890 \\ 18420 \\ 5690 \\ 10360 \\ 7385 \\ 14115 \\ 20545 \\ 9860 \\ 9910 \\ 11120 \\ 7275 \\ 22155 \\ 13260 \\ 19295 \\ 8535 \\ 6275 \\ 20855 \\ 6925 \\ 5660 \\ 8390 \\ 5675 \\ 3670 \\ 6200 \\ 7340 \\ 14325 \\ 4335 \\ 19815 \\ 9760 \\ 3800 \\ 12275 \\ 6320 \\ 3900 \\ 3345 \\ 6315 \\ 6035 \\ 6565 \\ 3310 \\ 5150 \\ 12710 \\ 4620 \\ 5775 \\ 3305 \\ 12430 \\ 14670 \\ 19905 \\ 4505 \\ 8030 \\ 6545 \\ 4965 \\ 5520 \\ 6225 \\ 5470 \\ 16120 \\ 4535 \\ 13865 \\ 10695 \\ 17455 \\ 4130 \\ 4130 \\ 4350 \\ 7200 \\ 11495 \\ 10865 \\ 4285 \\ 22660 \\ 5515 \\ 5835 \\ 7265 \\ 19705 \\ 4930 \\ 19125 \\ 6325 \\ 22020 \\ 4235 \\ 11250 \\ 7130 \\ 8115 \\ 13310 \\ 11115 \\ 16685 \\ 10510 \\ 10195 \\ 4265 \\ 6915 \\ 7465 \\ 11245 \\ 5220 \\ 7675 \\ 3785 \\ 23370 \\ 4370 \\ 6640 \\ 3690 \\ 16425 \\ 4065 \\ 11410 \\ 17085 \\ 8765 \\ 4240 \\ 3330 \\ 21910 \\ 16465 \\ 3480 \\ 3975 \\ 3930 \\ 5755 \\ 3155 \\ 9845 \\ 4290 \\ 11120 \\ 5715 \\ 9680 \\ 5895 \\ 5860 \\ 6210 \\ 12435 \\ 10850 \\ 4825 \\ 21820 \\ 14700 \\ 17065 \\ 8690 \\ 5280 \\ 4175 \\ 8125 \\ 6230 \\ 2760 \\ 13145 \\ 9140 \\ 4030 \\ 14435 \\ 17215 \\ 12780 \\ 9970 \\ 13470 \\ 4290 \\ 7565 \\ 10055 \\ 4600 \\ 14780 \\ 9465 \\ 14445 \\ 20920 \\ 7815 \\ 5465 \\ 9505 \\ 11475 \\ 9460 \\ 4975 \\ 11365 \\ 13350 \\ 9795 \\ 6160 \\ 4280 \\ 21820 \\ 9260 \\ 14735 \\ 3205 \\ 5220 \\ 22410 \\ 5770 \\ 20790 \\ 7025 \\ 4380 \\ 12105 \\ 4080 \\ 12070 \\ 6260 \\ 6995 \\ 7420 \\ 13185 \\ 16000 \\ 16500 \\ 19810 \\ 12810 \\ 13710 \\ 14570 \\ 5275 \\ 4010 \\ 8025 \\ 10275 \\ 6365 \\ 5405 \\ 5445 \\ 9820 \\ 12550 \\ 3635 \\ 8170 \\ 3855 \\ 6970 \\ 11075 \\ 3780 \\ 4405 \\ 6015 \\ 11900 \\ 4300 \\ 8230 \\ 7065 \\ 11170 \\ 5255 \\ 24725 \\ 8175 \\ 3545 \\ 11450 \\ 8840 \\ 3940 \\ 6590 \\ 6425 \\ 4130 \\ 9540 \\ 4390 \\ 8615 \\ 7885 \\ 4700 \\ 11475 \\ 6575 \\ 8945 \\ 9090 \\ 9490 \\ 5280 \\ 13405 \\ 7430 \\ 5985 \\ 5360 \\ 6430 \\ 10645 \\ 10535 \\ 6915 \\ 3790 \\ 22005 \\ 16935 \\ 11265 \\ 4580 \\ 4605 \\ 17675 \\ 13180 \\ 4460 \\ 5120 \\ 6840 \\ 4440 \\ 5760 \\ 12990 \\ 12195 \\ 13655 \\ 4880 \\ 13330 \\ 10820 \\ 13885 \\ 13670 \\ 6080 \\ 4365 \\ 6060 \\ 8050 \\ 3190 \\ 7150 \\ 17070 \\ 6470 \\ 22515 \\ 14275 \\ 3900 \\ 5740 \\ 10460 \\ 10885 \\ 10495 \\ 5120 \\ 4735 \\ 7945 \\ 5255 \\ 22720 \\ 16870 \\ 7760 \\ 12625 \\ 12050 \\ 11695 \\ 2955 \\ 13375 \\ 20970 \\ 4355 \\ 5330 \\ 5780 \\ 7215 \\ 10910 \\ 4730 \\ 7215 \\ 8810 \\ 7115 \\ 4350 \\ 8675 \\ 4805 \\ 4000 \\ 6725 \\ 6225 \\ 27010 \\ 6690 \\ 13915 \\ 25900 \\ 4980 \\ 13705 \\ 6655 \\ 6705 \\ 8150 \\ 2995 \\ 4945 \\ 5525 \\ 4825 \\ 13920 \\ 27540 \\ 4940 \\ 4825 \\ 5990 \\ 18860 \\ 5030 \\ 4775 \\ 9410 \\ 5650 \\ 4425 \\ 3785 \\ 7745 \\ 12500 \\ 6545 \\ 10660 \\ 5160 \\ 11105 \\ 8180 \\ 4035 \\ 4475 \\ 10005 \\ 19720 \\ 5505 \\ 25845 \\ 10735 \\ 8400 \\ 13440 \\ 8660 \\ 4345 \\ 4800 \\ 20755 \\ 9480 \\ 7980 \\ 9385 \\ 8795 \\ 8025 \\ 13960 \\ 5400 \\ 2890 \\ 10810 \\ 9985 \\ 19440 \\ 19705 \\ 4380 \\ 9000 \\ 12190 \\ 14755 \\ 6465 \\ 7080 \\ 6170 \\ 15675 \\ 3060 \\ 15525 \\ 8045 \\ 4995 \\ 4400 \\ 13370 \\ 17315 \\ 9615 \\ 2905 \\ 5660 \\ 5130 \\ 3455 \\ 5085 \\ 4470 \\ 3710 \\ 5035 \\ 3880 \\ 6645 \\ 12275 \\ 12830 \\ 9810 \\ 5095 \\ 15120 \\ 4370 \\ 7375 \\ 6105 \\ 7225 \\ 8685 \\ 8500 \\ 2995 \\ 5510 \\ 8110 \\ 2795 \\ 4890 \\ 15240 \\ 6910 \\ 4685 \\ 8140 \\ 8845 \\ 10615 \\ 7790 \\ 13280 \\ 12525 \\ 10520 \\ 6705 \\ 6615 \\ 3720 \\ 4920 \\ 10080 \\ 4515 \\ 4055 \\ 6825 \\ 5970 \\ 5305 \\ 9110 \\ 3660 \\ 20865 \\ 4030 \\ 8380 \\ 4335 \\ 7785 \\ 9685 \\ 7900 \\ 18130 \\ 8610 \\ 3040 \\ 23295 \\ 11960 \\ 3810 \\ 8070 \\ 10900 \\ 10000 \\ 13290 \\ 8495 \\ 3870 \\ 3125 \\ 9380 \\ 24365 \\ 3535 \\ 11185 \\ 11585 \\ 4235 \\ 8090 \\ 6820 \\ 7000 \\ 16305 \\ 11105 \\ 2605 \\ 11510 \\ 10965 \\ 8890 \\ 7460 \\ 6620 \\ 4440 \\ 3270 \\ 25445 \\ 4800 \\ 12290 \\ 9690 \\ 6170 \\ 5715 \\ 7315 \\ 7315 \\ 9685 \\ 9050 \\ 6145 \\ 8760 \\ 13240 \\ 4135 \\ 4375 \\ 7225 \\ 7460 \\ 3640 \\ 4130 \\ 11060 \\ 7980 \\ 5255 \\ 12085 \\ 11880 \\ 6090 \\ 25530 \\ 5710 \\ 3455 \\ 6955 \\ 6040 \\ 5320 \\ 8580 \\ 8800 \\ 9750 \\ 6300 \\ 7235 \\ 20030 \\ 5700 \\ 7395 \\ 11755 \\ 8655 \\ 2460 \\ 6710 \\ 12275 \\ 8300 \\ 6630 \\ 17390 \\ 13040 \\ 17910 \\ 4355 \\ 3745 \\ 7315 \\ 11115 \\ 17420 \\ 16655 \\ 7065 \\ 15235 \\ 6915 \\ 5100 \\ 5615 \\ 9365 \\ 3660 \\ 8585 \\ 12715 \\ 7845 \\ 9620 \\ 4515 \\ 4945 \\ 8865 \\ 6285 \\ 7935 \\ 8710 \\ 5485 \\ 7350 \\ 18000 \\ 6850 \\ 15905 \\ 15345 \\ 11760 \\ 9130 \\ 6875 \\ 5345 \\ 2825 \\ 12160 \\ 8265 \\ 15065 \\ 14960 \\ 15315 \\ 11615 \\ 2945 \\ 8995 \\ 4865 \\ 8685 \\ 4250 \\ 7375 \\ 7515 \\ 5655 \\ 6335 \\ 2990 \\ 4500 \\ 7915 \\ 3325 \\ 5365 \\ 15295 \\ 8555 \\ 5585 \\ 6990 \\ 10090 \\ 4825 \\ 4975 \\ 6535 \\ 25510 \\ 3925 \\ 20665 \\ 7725 \\ 8055 \\ 19220 \\ 15940 \\ 4065 \\ 5110 \\ 5470 \\ 25770 \\ 11675 \\ 3505 \\ 9400 \\ 7515 \\ 4755 \\ 13100 \\ 5090 \\ 8015 \\ 7910 \\ 9060 \\ 7800 \\ 28275 \\ 12230 \\ 11075 \\ 13110 \\ 13615 \\ 6640 \\ 6115 \\ 2870 \\ 23810 \\ 9165 \\ 18620 \\ 8020 \\ 17405 \\ 20885 \\ 5630 \\ 5655 \\ 10615 \\ 9670 \\ 10490 \\ 11305 \\ 6755 \\ 8350 \\ 12470 \\ 3505 \\ 15235 \\ 27345 \\ 3605 \\ 14245 \\ 26640 \\ 3710 \\ 6070 \\ 8310 \\ 11165 \\ 13745 \\ 9035 \\ 14410 \\
    };
			\addplot+[boxplot,black,style={solid},mark options={fill=black},mark=asterisk]
			table[row sep=\\,y index=0] {
			data\\
5385 \\ 2155 \\ 3690 \\ 1155 \\ 2265 \\ 4395 \\ 3590 \\ 1205 \\ 2565 \\ 4410 \\ 10225 \\ 3450 \\ 1735 \\ 3530 \\ 1625 \\ 4280 \\ 4515 \\ 3225 \\ 2105 \\ 1740 \\ 2275 \\ 2965 \\ 4340 \\ 2625 \\ 2090 \\ 1705 \\ 3575 \\ 7135 \\ 2980 \\ 3255 \\ 5840 \\ 1840 \\ 6645 \\ 2060 \\ 1555 \\ 1640 \\ 2025 \\ 2235 \\ 1090 \\ 2385 \\ 3535 \\ 7050 \\ 2010 \\ 4330 \\ 3865 \\ 2765 \\ 2570 \\ 3150 \\ 2340 \\ 5880 \\ 3085 \\ 6480 \\ 2205 \\ 2250 \\ 3005 \\ 3170 \\ 3300 \\ 1435 \\ 2530 \\ 2995 \\ 1220 \\ 5685 \\ 2465 \\ 1400 \\ 3665 \\ 2675 \\ 3050 \\ 1665 \\ 2290 \\ 3710 \\ 1920 \\ 3570 \\ 1605 \\ 2475 \\ 1120 \\ 2345 \\ 2765 \\ 1435 \\ 2405 \\ 2725 \\ 2280 \\ 5495 \\ 4590 \\ 2885 \\ 2895 \\ 1035 \\ 2330 \\ 1890 \\ 1155 \\ 2415 \\ 3055 \\ 2240 \\ 3185 \\ 1130 \\ 6460 \\ 2855 \\ 2055 \\ 2295 \\ 3200 \\ 1735 \\ 2400 \\ 3025 \\ 3385 \\ 5560 \\ 2275 \\ 5715 \\ 3385 \\ 4235 \\ 1820 \\ 2425 \\ 6715 \\ 2930 \\ 2345 \\ 2760 \\ 2590 \\ 3130 \\ 2280 \\ 2055 \\ 5150 \\ 2380 \\ 3475 \\ 4640 \\ 2355 \\ 2090 \\ 8720 \\ 4010 \\ 3285 \\ 2070 \\ 1070 \\ 1905 \\ 1605 \\ 1645 \\ 5680 \\ 2715 \\ 5450 \\ 1900 \\ 4755 \\ 5725 \\ 4505 \\ 2040 \\ 1645 \\ 2580 \\ 1625 \\ 3585 \\ 10610 \\ 2625 \\ 3645 \\ 2330 \\ 2620 \\ 4865 \\ 5065 \\ 1720 \\ 2845 \\ 2030 \\ 2405 \\ 4625 \\ 4655 \\ 5230 \\ 6880 \\ 2495 \\ 2220 \\ 4605 \\ 3650 \\ 3320 \\ 2720 \\ 3685 \\ 1265 \\ 6090 \\ 4330 \\ 6430 \\ 3370 \\ 6785 \\ 1735 \\ 2710 \\ 2505 \\ 1585 \\ 2610 \\ 2100 \\ 6935 \\ 4060 \\ 7465 \\ 2460 \\ 3150 \\ 2445 \\ 2500 \\ 3810 \\ 7985 \\ 3645 \\ 4095 \\ 3430 \\ 1155 \\ 4815 \\ 2245 \\ 5735 \\ 2610 \\ 4285 \\ 3185 \\ 3940 \\ 3980 \\ 3105 \\ 3095 \\ 2100 \\ 2720 \\ 2010 \\ 1985 \\ 1605 \\ 6040 \\ 2300 \\ 5055 \\ 2380 \\ 2180 \\ 3825 \\ 5850 \\ 3305 \\ 6920 \\ 2380 \\ 2650 \\ 5420 \\ 2815 \\ 1575 \\ 3625 \\ 2865 \\ 2025 \\ 5475 \\ 4350 \\ 1180 \\ 3875 \\ 3610 \\ 2240 \\ 1985 \\ 1955 \\ 3680 \\ 2965 \\ 4135 \\ 2610 \\ 1130 \\ 4345 \\ 6265 \\ 2315 \\ 4225 \\ 5670 \\ 3785 \\ 1635 \\ 3205 \\ 1900 \\ 2065 \\ 2390 \\ 2310 \\ 3195 \\ 2800 \\ 5245 \\ 3890 \\ 3180 \\ 2435 \\ 2525 \\ 3140 \\ 2320 \\ 2845 \\ 4790 \\ 2030 \\ 3520 \\ 3410 \\ 1130 \\ 1985 \\ 7450 \\ 2480 \\ 2885 \\ 2170 \\ 2700 \\ 1265 \\ 3865 \\ 4360 \\ 1545 \\ 4015 \\ 2005 \\ 6020 \\ 3135 \\ 2645 \\ 4330 \\ 3460 \\ 7675 \\ 3300 \\ 1840 \\ 3750 \\ 1525 \\ 2795 \\ 7125 \\ 3375 \\ 2205 \\ 4245 \\ 2550 \\ 3135 \\ 5470 \\ 2815 \\ 3690 \\ 2885 \\ 2025 \\ 1875 \\ 2980 \\ 3035 \\ 2420 \\ 3095 \\ 2420 \\ 1555 \\ 1620 \\ 2735 \\ 3290 \\ 2390 \\ 3255 \\ 2160 \\ 4480 \\ 2515 \\ 4115 \\ 2945 \\ 4315 \\ 2415 \\ 7950 \\ 5420 \\ 3705 \\ 2960 \\ 2130 \\ 2095 \\ 1265 \\ 4540 \\ 7750 \\ 1160 \\ 5905 \\ 3445 \\ 1475 \\ 2810 \\ 1605 \\ 5290 \\ 5450 \\ 1620 \\ 1120 \\ 4245 \\ 1665 \\ 2410 \\ 2780 \\ 3725 \\ 3790 \\ 3070 \\ 4500 \\ 1605 \\ 2320 \\ 4400 \\ 3125 \\ 1620 \\ 2000 \\ 3715 \\ 4130 \\ 3280 \\ 2455 \\ 2650 \\ 3650 \\ 2255 \\ 1370 \\ 3305 \\ 1575 \\ 1465 \\ 4395 \\ 1540 \\ 1935 \\ 3310 \\ 3005 \\ 3085 \\ 3260 \\ 3305 \\ 2440 \\ 4965 \\ 1535 \\ 4695 \\ 3020 \\ 1225 \\ 3400 \\ 5030 \\ 3115 \\ 4995 \\ 1615 \\ 2775 \\ 3235 \\ 3750 \\ 1305 \\ 2845 \\ 6060 \\ 2755 \\ 2350 \\ 4405 \\ 3230 \\ 3010 \\ 3700 \\ 2710 \\ 2360 \\ 3555 \\ 1380 \\ 1225 \\ 6185 \\ 2605 \\ 2835 \\ 2095 \\ 2120 \\ 3365 \\ 2040 \\ 1965 \\ 1815 \\ 1785 \\ 3260 \\ 1665 \\ 2185 \\ 4495 \\ 3940 \\ 2455 \\ 2330 \\ 1160 \\ 4385 \\ 8270 \\ 1090 \\ 3930 \\ 1785 \\ 3490 \\ 2145 \\ 2400 \\ 2535 \\ 3375 \\ 2290 \\ 2340 \\ 2880 \\ 5665 \\ 1630 \\ 5050 \\ 2075 \\ 1910 \\ 2305 \\ 6440 \\ 3085 \\ 2085 \\ 3320 \\ 5880 \\ 2245 \\ 2435 \\ 2760 \\ 2315 \\ 2315 \\ 2360 \\ 3220 \\ 1650 \\ 2740 \\ 3515 \\ 2965 \\ 1990 \\ 3030 \\ 4170 \\ 3655 \\ 3025 \\ 4280 \\ 1045 \\ 2825 \\ 5040 \\ 5915 \\ 1160 \\ 2455 \\ 2045 \\ 2570 \\ 1535 \\ 2325 \\ 5710 \\ 1765 \\ 2025 \\ 3780 \\ 2540 \\ 2145 \\ 2065 \\ 2645 \\ 1135 \\ 7335 \\ 1055 \\ 3510 \\ 1155 \\ 4945 \\ 2810 \\ 2615 \\ 5860 \\ 5150 \\ 5240 \\ 2275 \\ 7250 \\ 1935 \\ 1550 \\ 5485 \\ 1175 \\ 1620 \\ 7420 \\ 1905 \\ 3035 \\ 2205 \\ 1205 \\ 2600 \\ 2540 \\ 3225 \\ 5760 \\ 3495 \\ 3410 \\ 1605 \\ 3455 \\ 2205 \\ 2655 \\ 2835 \\ 4465 \\ 2595 \\ 5150 \\ 3565 \\ 4370 \\ 1935 \\ 2495 \\ 3630 \\ 2890 \\ 1565 \\ 2105 \\ 2080 \\ 5520 \\ 2945 \\ 1645 \\ 3950 \\ 5140 \\ 6220 \\ 5550 \\ 1295 \\ 2610 \\ 2880 \\ 1865 \\ 1200 \\ 1445 \\ 5350 \\ 1210 \\ 1960 \\ 5050 \\ 2020 \\ 5410 \\ 2765 \\ 1205 \\ 2305 \\ 5790 \\ 3690 \\ 2800 \\ 5015 \\ 1045 \\ 2605 \\ 2380 \\ 1280 \\ 2820 \\ 2530 \\ 3750 \\ 2850 \\ 4235 \\ 5240 \\ 2805 \\ 6625 \\ 3385 \\ 2305 \\ 1755 \\ 8040 \\ 2740 \\ 4400 \\ 2625 \\ 1170 \\ 5860 \\ 7440 \\ 3625 \\ 6130 \\ 4490 \\ 1610 \\ 3055 \\ 2145 \\ 4205 \\ 4645 \\ 2895 \\ 1705 \\ 2835 \\ 1105 \\ 1895 \\ 2950 \\ 2035 \\ 5315 \\ 3990 \\ 3905 \\ 1060 \\ 2140 \\ 2585 \\ 2450 \\ 1655 \\ 3670 \\ 5690 \\ 2680 \\ 2215 \\ 1925 \\ 4685 \\ 3315 \\ 4905 \\ 3295 \\ 3455 \\ 1760 \\ 4195 \\ 4195 \\ 1865 \\ 2290 \\ 3465 \\ 2235 \\ 1400 \\ 3305 \\ 4355 \\ 3930 \\ 3475 \\ 2895 \\ 3575 \\ 2020 \\ 2585 \\ 3100 \\ 1265 \\ 1235 \\ 2415 \\ 3760 \\ 3290 \\ 1790 \\ 3585 \\ 1665 \\ 2600 \\ 7795 \\ 2580 \\ 2210 \\ 2015 \\ 2560 \\ 1925 \\ 2690 \\ 1525 \\ 2590 \\ 2520 \\ 5945 \\ 2415 \\ 2095 \\ 7525 \\ 5050 \\ 2145 \\ 3930 \\ 4860 \\ 4655 \\ 2645 \\ 1805 \\ 5715 \\ 3115 \\ 8215 \\ 3980 \\ 3230 \\ 2185 \\ 3270 \\ 4555 \\ 2425 \\ 4735 \\ 2180 \\ 2585 \\ 2155 \\ 3230 \\ 2800 \\ 5830 \\ 1440 \\ 1805 \\ 1330 \\ 2205 \\ 3850 \\ 2810 \\ 4635 \\ 2080 \\ 3255 \\ 2095 \\ 2080 \\ 3190 \\ 3345 \\ 2745 \\ 3890 \\ 2845 \\ 6785 \\ 2395 \\ 2975 \\ 3325 \\ 1060 \\ 2900 \\ 3185 \\ 3595 \\ 4040 \\ 2340 \\ 2185 \\ 2255 \\ 4290 \\ 1685 \\ 5085 \\ 10425 \\ 3340 \\ 3730 \\ 5485 \\ 2380 \\ 2095 \\ 3925 \\ 3725 \\ 1505 \\ 2270 \\ 1575 \\ 3060 \\ 2205 \\ 3115 \\ 2060 \\ 1635 \\ 2210 \\ 2805 \\ 3900 \\ 2605 \\ 2905 \\ 3545 \\ 3820 \\ 4100 \\ 4665 \\ 3300 \\ 2305 \\ 3395 \\ 2305 \\ 7075 \\ 5685 \\ 2935 \\ 3090 \\ 1935 \\ 1410 \\ 8045 \\ 6810 \\ 1845 \\ 1850 \\ 4210 \\ 3270 \\ 3255 \\ 2210 \\ 3730 \\ 3260 \\ 2485 \\ 3080 \\ 2630 \\ 3275 \\ 1665 \\ 3245 \\ 1405 \\ 2905 \\ 1225 \\ 5960 \\ 1900 \\ 2170 \\ 5920 \\ 3005 \\ 2925 \\ 3265 \\ 2960 \\ 3135 \\ 8635 \\ 1405 \\ 1590 \\ 2185 \\ 4485 \\ 2335 \\ 2865 \\ 2045 \\ 1725 \\ 2375 \\ 1930 \\ 2010 \\ 2220 \\ 1415 \\ 2345 \\ 2490 \\ 6250 \\ 2325 \\ 2090 \\ 3840 \\ 5450 \\ 2525 \\ 3040 \\ 2435 \\ 2485 \\ 1655 \\ 4170 \\ 2130 \\ 1975 \\ 3570 \\ 1470 \\ 4210 \\ 5640 \\ 5250 \\ 2750 \\ 11425 \\ 3395 \\ 2080 \\ 1070 \\ 2655 \\ 4520 \\ 3935 \\ 3395 \\ 5525 \\ 2025 \\ 3315 \\ 1165 \\ 3240 \\ 1375 \\ 3445 \\ 4755 \\ 3080 \\ 4040 \\ 2035 \\ 2145 \\ 1745 \\ 4055 \\ 3185 \\ 1160 \\ 4935 \\ 1150 \\ 1435 \\ 3025 \\ 4315 \\ 2395 \\ 2820 \\ 2970 \\ 4450 \\ 2700 \\ 3125 \\ 1490 \\ 2140 \\ 4630 \\ 3745 \\ 1165 \\ 1660 \\ 5660 \\ 2420 \\ 2320 \\ 2725 \\ 2665 \\ 3210 \\ 3770 \\ 5015 \\ 4405 \\ 2950 \\ 1065 \\ 1690 \\ 1165 \\ 2255 \\ 7880 \\ 2680 \\ 3525 \\ 4380 \\ 2055 \\ 3165 \\ 2795 \\ 2815 \\ 3905 \\ 3790 \\ 4355 \\ 7295 \\ 2225 \\ 3815 \\ 1180 \\ 3890 \\ 1175 \\ 1870 \\ 8180 \\ 5995 \\ 3780 \\ 2200 \\ 2175 \\ 3545 \\ 4230 \\ 3900 \\ 3080 \\ 2395 \\ 2680 \\ 3385 \\ 1130 \\ 2455 \\ 2735 \\ 1620 \\ 3025 \\ 6545 \\ 2820 \\ 1675 \\ 1575 \\ 2800 \\ 2750 \\ 4025 \\ 2620 \\ 1665 \\ 2240 \\ 1675 \\ 3385 \\ 3630 \\ 2825 \\ 2685 \\ 2105 \\ 2050 \\ 3700 \\ 5370 \\ 2430 \\ 3030 \\ 6765 \\ 2720 \\ 7745 \\ 1665 \\ 1895 \\ 3350 \\ 1935 \\ 3835 \\ 8125 \\ 2220 \\ 4395 \\ 2750 \\ 3010 \\ 7580 \\ 1625 \\ 2915 \\ 2005 \\ 2770 \\ 2230 \\ 5975 \\ 4340 \\ 3045 \\ 2500 \\ 11580 \\ 2015 \\ 3125 \\ 3890 \\ 1110 \\ 1670 \\ 1560 \\ 1990 \\ 4580 \\ 4585 \\ 1270 \\ 7105 \\ 1120 \\ 3295 \\ 2365 \\ 4770 \\ 3675 \\ 3735 \\ 1530 \\ 2465 \\ 3495 \\ 1165 \\ 2370 \\ 3940 \\ 1995 \\ 2000 \\ 6465 \\ 1945 \\ 2540 \\ 2460 \\ 4440 \\ 4730 \\ 3620 \\ 1215 \\ 3180 \\ 2025 \\ 3120 \\ 3805 \\ 1630 \\ 2405 \\ 1465 \\ 4185 \\ 2330 \\ 3060 \\ 4975 \\ 1060 \\ 1735 \\ 7495 \\ 2215 \\ 4480 \\ 4830 \\ 1380 \\ 4365 \\ 4485 \\ 4175 \\ 1250 \\ 2885 \\ 4015 \\ 2905 \\ 7090 \\ 1055 \\ 2010 \\ 2970 \\
    };
			\addplot+[boxplot,black,style={solid},mark options={fill=black},mark=asterisk]
			table[row sep=\\,y index=0] {
			data\\
1880 \\ 1925 \\ 4230 \\ 3235 \\ 2625 \\ 2700 \\ 5840 \\ 2445 \\ 8185 \\ 6120 \\ 3960 \\ 4635 \\ 2885 \\ 6745 \\ 2940 \\ 2175 \\ 2830 \\ 2230 \\ 2710 \\ 3275 \\ 3995 \\ 1905 \\ 1935 \\ 2460 \\ 7880 \\ 2385 \\ 2385 \\ 2405 \\ 1810 \\ 3165 \\ 9000 \\ 4455 \\ 2740 \\ 5860 \\ 2090 \\ 6505 \\ 2990 \\ 2830 \\ 4430 \\ 3655 \\ 2185 \\ 4485 \\ 3690 \\ 3880 \\ 2020 \\ 3755 \\ 1805 \\ 2045 \\ 6975 \\ 4650 \\ 1915 \\ 4610 \\ 2900 \\ 2030 \\ 2445 \\ 4945 \\ 2045 \\ 2780 \\ 1860 \\ 6010 \\ 3940 \\ 4695 \\ 5195 \\ 1980 \\ 2685 \\ 7020 \\ 2650 \\ 1850 \\ 3575 \\ 4290 \\ 2740 \\ 22620 \\ 6475 \\ 2785 \\ 2165 \\ 3345 \\ 4185 \\ 2425 \\ 1840 \\ 5300 \\ 1800 \\ 2980 \\ 3265 \\ 2275 \\ 2060 \\ 5510 \\ 1915 \\ 7330 \\ 2165 \\ 2260 \\ 4405 \\ 2700 \\ 4435 \\ 1905 \\ 2000 \\ 6630 \\ 4305 \\ 5260 \\ 1815 \\ 3570 \\ 3705 \\ 4195 \\ 2715 \\ 4775 \\ 3560 \\ 2575 \\ 2345 \\ 2840 \\ 4505 \\ 2240 \\ 4140 \\ 4245 \\ 9450 \\ 1925 \\ 2885 \\ 2965 \\ 5025 \\ 1800 \\ 8555 \\ 2250 \\ 3030 \\ 2810 \\ 4975 \\ 2280 \\ 2125 \\ 4895 \\ 1885 \\ 9780 \\ 2725 \\ 4935 \\ 1835 \\ 2440 \\ 2530 \\ 3790 \\ 8730 \\ 2745 \\ 3225 \\ 2890 \\ 3300 \\ 1830 \\ 5970 \\ 3630 \\ 3325 \\ 1820 \\ 2830 \\ 7415 \\ 3110 \\ 2305 \\ 2255 \\ 1825 \\ 9270 \\ 5505 \\ 1805 \\ 7930 \\ 1875 \\ 3145 \\ 2605 \\ 3650 \\ 3710 \\ 1960 \\ 4625 \\ 1880 \\ 1995 \\ 9070 \\ 2225 \\ 2865 \\ 3020 \\ 4850 \\ 2235 \\ 4710 \\ 5320 \\ 1890 \\ 8905 \\ 4825 \\ 2080 \\ 4375 \\ 3360 \\ 2315 \\ 4980 \\ 1900 \\ 1775 \\ 10275 \\ 1890 \\ 2210 \\ 3420 \\ 3140 \\ 1840 \\ 6760 \\ 1945 \\ 1955 \\ 1825 \\ 1835 \\ 2315 \\ 7250 \\ 3110 \\ 2980 \\ 1955 \\ 3345 \\ 3560 \\ 2795 \\ 7100 \\ 2050 \\ 6715 \\ 1820 \\ 2815 \\ 4180 \\ 4375 \\ 2510 \\ 1825 \\ 1880 \\ 2355 \\ 6970 \\ 2865 \\ 5995 \\ 4310 \\ 7290 \\ 2535 \\ 2145 \\ 6605 \\ 6275 \\ 4100 \\ 3000 \\ 1825 \\ 8780 \\ 6100 \\ 10500 \\ 2475 \\ 1840 \\ 1990 \\ 2920 \\ 1950 \\ 4430 \\ 2910 \\ 2080 \\ 6750 \\ 3230 \\ 1875 \\ 2880 \\ 4050 \\ 1915 \\ 2400 \\ 2950 \\ 5090 \\ 7625 \\ 2155 \\ 2750 \\ 5800 \\ 2605 \\ 5410 \\ 2770 \\ 4135 \\ 2725 \\ 2255 \\ 2900 \\ 1835 \\ 1920 \\ 4110 \\ 2785 \\ 7610 \\ 1795 \\ 7145 \\ 3970 \\ 6285 \\ 15870 \\ 9585 \\ 3970 \\ 3380 \\ 2685 \\ 1915 \\ 1840 \\ 2150 \\ 2775 \\ 2510 \\ 1810 \\ 7755 \\ 5810 \\ 4340 \\ 6470 \\ 1915 \\ 4620 \\ 5015 \\ 2480 \\ 1985 \\ 2095 \\ 3105 \\ 3410 \\ 5600 \\ 3750 \\ 8180 \\ 2875 \\ 1830 \\ 2875 \\ 2790 \\ 6465 \\ 3190 \\ 2670 \\ 2680 \\ 5960 \\ 2590 \\ 3845 \\ 3710 \\ 2330 \\ 2995 \\ 1920 \\ 6785 \\ 2405 \\ 5480 \\ 4910 \\ 12800 \\ 2350 \\ 1810 \\ 6590 \\ 2870 \\ 5565 \\ 2665 \\ 2915 \\ 1865 \\ 5055 \\ 3410 \\ 3130 \\ 2330 \\ 2390 \\ 1805 \\ 2310 \\ 2575 \\ 2945 \\ 2255 \\ 5480 \\ 1845 \\ 4495 \\ 7575 \\ 4215 \\ 3605 \\ 3520 \\ 3665 \\ 2930 \\ 2540 \\ 2835 \\ 3480 \\ 8385 \\ 5380 \\ 2720 \\ 3915 \\ 5705 \\ 2270 \\ 2075 \\ 2255 \\ 4320 \\ 5325 \\ 4515 \\ 2625 \\ 2770 \\ 4685 \\ 3030 \\ 14550 \\ 1980 \\ 4615 \\ 6755 \\ 1815 \\ 2410 \\ 4440 \\ 1800 \\ 2270 \\ 3585 \\ 3225 \\ 3620 \\ 2320 \\ 2025 \\ 2670 \\ 7815 \\ 1830 \\ 2975 \\ 1910 \\ 2395 \\ 3460 \\ 1930 \\ 10050 \\ 1870 \\ 4775 \\ 2080 \\ 2110 \\ 4070 \\ 11045 \\ 5805 \\ 2425 \\ 3210 \\ 3800 \\ 2990 \\ 3505 \\ 3260 \\ 5405 \\ 1840 \\ 5470 \\ 2080 \\ 1815 \\ 4035 \\ 2730 \\ 1825 \\ 1890 \\ 8490 \\ 2020 \\ 1820 \\ 2170 \\ 2000 \\ 3230 \\ 4955 \\ 1820 \\ 6580 \\ 2995 \\ 1830 \\ 4445 \\ 8960 \\ 3365 \\ 1800 \\ 6750 \\ 3685 \\ 6840 \\ 2645 \\ 10195 \\ 5150 \\ 4955 \\ 1870 \\ 2400 \\ 3985 \\ 7090 \\ 2535 \\ 3735 \\ 4670 \\ 3600 \\ 5200 \\ 9760 \\ 3675 \\ 3930 \\ 6480 \\ 4535 \\ 2560 \\ 1950 \\ 1840 \\ 2005 \\ 4735 \\ 3390 \\ 3330 \\ 3775 \\ 5790 \\ 3585 \\ 3605 \\ 7610 \\ 4075 \\ 4740 \\ 5520 \\ 2580 \\ 2235 \\ 1860 \\ 4560 \\ 8725 \\ 5035 \\ 2590 \\ 2330 \\ 3970 \\ 3850 \\ 2095 \\ 4970 \\ 3480 \\ 2315 \\ 2135 \\ 2610 \\ 2605 \\ 4225 \\ 5560 \\ 3240 \\ 3375 \\ 2270 \\ 2410 \\ 2425 \\ 5065 \\ 1925 \\ 2000 \\ 3130 \\ 1835 \\ 5540 \\ 1820 \\ 4340 \\ 4575 \\ 3625 \\ 2665 \\ 1830 \\ 3365 \\ 2355 \\ 4190 \\ 3945 \\ 5150 \\ 3165 \\ 1930 \\ 5650 \\ 1815 \\ 1870 \\ 4615 \\ 6710 \\ 2625 \\ 2320 \\ 9510 \\ 2720 \\ 6160 \\ 2885 \\ 3930 \\ 9465 \\ 1995 \\ 2440 \\ 6035 \\ 2560 \\ 5490 \\ 1920 \\ 1840 \\ 3630 \\ 1910 \\ 9290 \\ 2105 \\ 3425 \\ 3715 \\ 1840 \\ 2260 \\ 12160 \\ 2865 \\ 8270 \\ 2965 \\ 10460 \\ 3295 \\ 3345 \\ 1880 \\ 2750 \\ 1970 \\ 4020 \\ 4280 \\ 3055 \\ 2570 \\ 6200 \\ 1930 \\ 3040 \\ 3570 \\ 4555 \\ 4640 \\ 2825 \\ 4005 \\ 2905 \\ 8470 \\ 2165 \\ 2100 \\ 1820 \\ 3020 \\ 5020 \\ 1930 \\ 11200 \\ 2720 \\ 2205 \\ 1960 \\ 3030 \\ 1860 \\ 9540 \\ 10340 \\ 3375 \\ 1940 \\ 2125 \\ 3810 \\ 1815 \\ 2700 \\ 3350 \\ 3125 \\ 2645 \\ 5540 \\ 2510 \\ 3140 \\ 4760 \\ 1805 \\ 3980 \\ 2140 \\ 1875 \\ 2815 \\ 1800 \\ 2310 \\ 3210 \\ 5555 \\ 2830 \\ 2630 \\ 3240 \\ 4540 \\ 2270 \\ 3360 \\ 1920 \\ 5560 \\ 5955 \\ 1820 \\ 1815 \\ 3180 \\ 2340 \\ 2885 \\ 2540 \\ 2120 \\ 1915 \\ 3075 \\ 2840 \\ 3020 \\ 5720 \\ 2455 \\ 2025 \\ 2495 \\ 3505 \\ 3065 \\ 4035 \\ 1955 \\ 9625 \\ 1890 \\ 2845 \\ 7220 \\ 3040 \\ 4090 \\ 7035 \\ 5595 \\ 4725 \\ 2410 \\ 1940 \\ 1940 \\ 3955 \\ 7080 \\ 1955 \\ 2360 \\ 2855 \\ 1875 \\ 2380 \\ 4790 \\ 4265 \\ 2195 \\ 1830 \\ 12665 \\ 1870 \\ 8140 \\ 9765 \\ 3235 \\ 3355 \\ 1800 \\ 4165 \\ 4460 \\ 3140 \\ 3515 \\ 4755 \\ 2055 \\ 3575 \\ 3190 \\ 5120 \\ 4615 \\ 3200 \\ 3655 \\ 2505 \\ 1815 \\ 6870 \\ 4425 \\ 2190 \\ 4790 \\ 6690 \\ 4955 \\ 2650 \\ 4900 \\ 3445 \\ 5265 \\ 1920 \\ 3035 \\ 2010 \\ 3900 \\ 1820 \\ 2305 \\ 5645 \\ 3625 \\ 6600 \\ 2770 \\ 2060 \\ 2315 \\ 2290 \\ 4335 \\ 1835 \\ 6075 \\ 5840 \\ 1810 \\ 3820 \\ 1860 \\ 3820 \\ 2330 \\ 5340 \\ 3885 \\ 3285 \\ 2550 \\ 3005 \\ 5240 \\ 6025 \\ 2695 \\ 5745 \\ 5590 \\ 2115 \\ 2200 \\ 2360 \\ 1960 \\ 3415 \\ 3590 \\ 1875 \\ 4000 \\ 4140 \\ 1995 \\ 5795 \\ 2060 \\ 7480 \\ 2485 \\ 1840 \\ 2625 \\ 5630 \\ 1890 \\ 4800 \\ 1820 \\ 2875 \\ 3060 \\ 2160 \\ 4710 \\ 1845 \\ 4845 \\ 6500 \\ 2060 \\ 3705 \\ 1810 \\ 1820 \\ 1905 \\ 4915 \\ 6965 \\ 8450 \\ 2835 \\ 3985 \\ 4210 \\ 1880 \\ 5565 \\ 7215 \\ 1935 \\ 10800 \\ 2250 \\ 1775 \\ 1805 \\ 4610 \\ 2950 \\ 4790 \\ 2870 \\ 6420 \\ 6425 \\ 5530 \\ 2295 \\ 2965 \\ 2635 \\ 3935 \\ 3030 \\ 5060 \\ 13780 \\ 2735 \\ 2485 \\ 5375 \\ 1925 \\ 2325 \\ 3515 \\ 2230 \\ 2200 \\ 6080 \\ 5495 \\ 2895 \\ 2980 \\ 4630 \\ 2000 \\ 4585 \\ 2385 \\ 6140 \\ 1860 \\ 3570 \\ 2520 \\ 2635 \\ 4120 \\ 1990 \\ 7535 \\ 1835 \\ 1975 \\ 2745 \\ 2605 \\ 5805 \\ 1810 \\ 5045 \\ 3695 \\ 2685 \\ 1930 \\ 3030 \\ 2750 \\ 2640 \\ 1865 \\ 2795 \\ 2980 \\ 4310 \\ 3980 \\ 4850 \\ 3015 \\ 2685 \\ 2400 \\ 3865 \\ 2925 \\ 4075 \\ 2270 \\ 2400 \\ 8190 \\ 2110 \\ 6135 \\ 5145 \\ 3820 \\ 3965 \\ 2735 \\ 3590 \\ 2280 \\ 1895 \\ 4685 \\ 2255 \\ 4715 \\ 4420 \\ 2415 \\ 1915 \\ 4345 \\ 2240 \\ 5840 \\ 1845 \\ 3770 \\ 2880 \\ 2730 \\ 4515 \\ 3270 \\ 2715 \\ 2815 \\ 3370 \\ 1840 \\ 3350 \\ 2025 \\ 1860 \\ 2685 \\ 3645 \\ 6140 \\ 7905 \\ 2995 \\ 2520 \\ 2685 \\ 3150 \\ 4125 \\ 1950 \\ 2355 \\ 2360 \\ 2950 \\ 4745 \\ 2735 \\ 4640 \\ 2850 \\ 2745 \\ 5230 \\ 12010 \\ 1885 \\ 2180 \\ 2075 \\ 3080 \\ 6750 \\ 2895 \\ 1815 \\ 2710 \\ 2260 \\ 7265 \\ 2055 \\ 2665 \\ 3070 \\ 1885 \\ 2780 \\ 1915 \\ 1850 \\ 6565 \\ 3065 \\ 1875 \\ 7210 \\ 3265 \\ 2495 \\ 3605 \\ 2720 \\ 2620 \\ 3765 \\ 3315 \\ 4030 \\ 2805 \\ 2485 \\ 8845 \\ 2335 \\ 1895 \\ 2730 \\ 3940 \\ 5570 \\ 2940 \\ 6075 \\ 2135 \\ 5335 \\ 2765 \\ 4705 \\ 2750 \\ 10385 \\ 5435 \\ 6300 \\ 2470 \\ 2205 \\ 3250 \\ 2240 \\ 9160 \\ 3350 \\ 6105 \\ 4360 \\ 3070 \\ 2370 \\ 4505 \\ 2040 \\ 2850 \\ 2970 \\ 4680 \\ 1850 \\ 3675 \\ 1920 \\ 7040 \\ 2715 \\ 2115 \\ 4125 \\ 5310 \\ 3480 \\ 2415 \\ 1875 \\ 3365 \\ 1815 \\ 1950 \\ 2855 \\ 9220 \\ 2665 \\ 1825 \\ 1910 \\ 2520 \\ 1840 \\ 13205 \\ 1855 \\ 7080 \\ 5450 \\ 2055 \\ 3815 \\ 7010 \\ 2525 \\ 6805 \\ 3015 \\ 3765 \\ 2500 \\ 4055 \\ 2375 \\ 5610 \\ 6110 \\ 2785 \\ 1855 \\ 2170 \\ 3040 \\ 2910 \\ 2335 \\ 3125 \\ 4550 \\ 1960 \\ 6030 \\ 1825 \\ 6345 \\ 2695 \\ 1955 \\ 2555 \\ 3690 \\ 4190 \\ 2140 \\ 7165 \\ 5050 \\ 5265 \\ 3070 \\ 2115 \\ 8575 \\ 12205 \\ 2190 \\ 8970 \\ 3180 \\ 2890 \\ 2700 \\ 2280 \\ 2525 \\ 8460 \\ 1985 \\ 1940 \\ 3125 \\ 7065 \\
    };
		\end{axis}
	\end{tikzpicture}
	\caption{Distributions of solution times for different levels of cooperation in an environment containing two fixed barriers.}
	\label{fig:sts}
\end{figure}

A number of experiments have been conducted, each using multiple simulation runs to investigate different configurations of obstacles and levels of cooperation.  For example, \figurename~\ref{fig:sts} shows the distribution of solution times resulting from 1000 runs in an environment containing two fixed barriers (configured as shown in \figurename~\ref{fig:SimEnv}.) for four incremental levels of cooperation.  As expected, enabling explicit communication between the agents had a measurable effect in speeding up solution times.  However, it was also noted that the low solution times for [1000] was due to the high number of runs that did not finish (DNF).  In particular, the results revealed that [1000] gave rise to 64\% DNFs, whereas [1100] had 13\% DNFs, [1110] had 1\% DNFs, and [1111] had only 0.6\% DNFs.  

In attempting to analyse the results of the more complex cooperation experiments, it became clear that an overall `goodness measure' was needed in order to resolve the compromise between fast solution times and the numbers of runs that did not finish.  This was necessary because, as seen above, a high number of DNFs tends to give rise to a low mean solution time because the runs that succeed have less challenging barrier configurations.  Likewise, a low number of DNFs may be associated with higher mean solution times as a consequence of the cooperating agents taking longer to solve more challenging barrier configurations.

Hence, an appropriate `goodness measure' was defined as:
\begin{equation}
 GM = log(ST^{1+\frac{DNF}{nruns}}), 
\end{equation}
where $GM$ is the goodness measure (low is good), $ST$ is the mean solution time for a run, $DNF$ is the number of times a run did not finish, and $nruns$ is the number of runs.

\begin{figure}[h!]
	\centering
	\begin{tikzpicture}
		\begin{axis}[xlabel=Cooperation Level,ylabel={Goodness Measure, GM},xtick={1,2,3,4},xticklabels={[1000], [1100], [1110], [1111]}]
			\addplot+[black, only marks, mark options={fill=black}, error bars/.cd,
			y dir=both,y explicit]
			coordinates {
			(1, 6.0828)
			(2, 4.4972)
			(3, 3.5445)
			(4, 3.5989)
			};
		\end{axis}
	\end{tikzpicture}
	\caption{Relationship between the cooperation level and the `goodness measure' (low is good) in an environment containing two fixed barriers.}
	\label{fig:gm}
\end{figure}
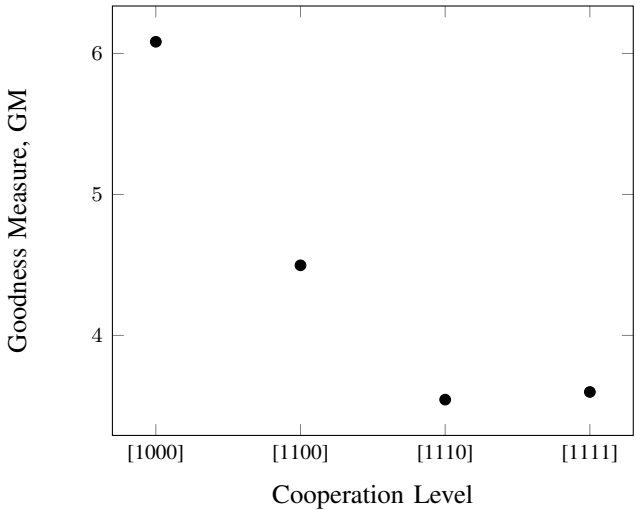

\figurename~\ref{fig:gm} shows the combined results from the solution times shown in \figurename~\ref{fig:sts} and the corresponding number of DNFs plotted using the goodness measure.  This representation clearly shows that, as expected, increasing the level of cooperation between the agents leads to significant improvements in their ability to solve the designated task.

\subsection{Matched Agents}

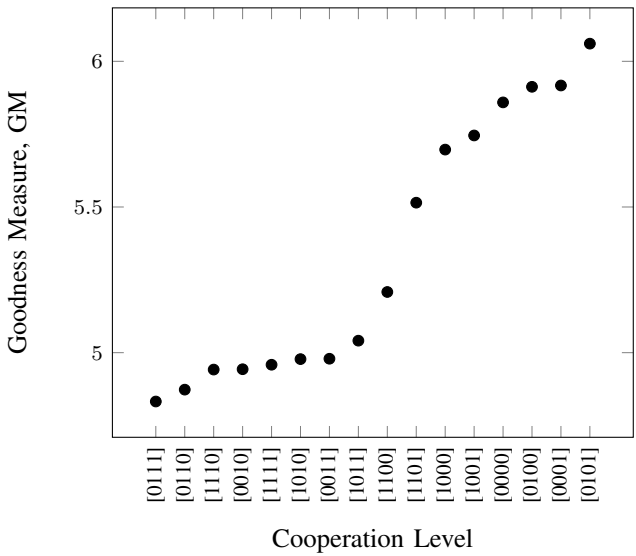
\begin{figure}[h!]
	\centering
	\begin{tikzpicture}
		\begin{axis}[x label style={at={(axis description cs:0.5,-0.1)},anchor=north},xlabel=Cooperation Level, ylabel={Goodness Measure, GM},xtick={0, 1, 2, 3, 4, 5, 6, 7, 8, 9, 10, 11, 12, 13, 14, 15},xticklabels={[0111], [0110], [1110], [0010], [1111], [1010], [0011], [1011], [1100], [1101], [1000], [1001], [0000], [0100], [0001], [0101]},xticklabel style={rotate=90}]
			\addplot+[black, only marks, mark options={fill=black}, error bars/.cd,
			y dir=both,y explicit]
			coordinates {
			(12, 5.8590)
			(10, 5.6970)
			(13, 5.9125)
			(8, 5.2083)
			(3, 4.9432)
			(5, 4.9779)
			(1, 4.8731)
			(2, 4.9420)
			(14, 5.9170)
			(11, 5.7455)
			(15, 6.0606)
			(9, 5.5148)
			(6, 4.9792)
			(7, 5.0412)
			(0, 4.8324)
			(4, 4.9586)
			};
		\end{axis}
	\end{tikzpicture}
	\caption{Relationship between different cooperation combinations ordered by their `goodness' (low is good) for matched agents in an environment containing three randomly placed barriers with random lengths and orientations.}
	\label{fig:good}
\end{figure}

As mentioned above, the simulation environment allowed the cooperation level to be set for each agent independently.  However, due to the combinatorics, the majority of experiments were conducted with \emph{matched} agents.  For example, \figurename~\ref{fig:good} shows the impact of all sixteen levels of cooperation ranked by the `goodness' of the outcome for matched agents in an environment containing three randomly placed barriers with random lengths and orientations.

As can be seen in \figurename~\ref{fig:good}, the relationship between different combinations of cooperation and the goodness measure reveals that the cooperation combination [0111] ``\emph{arrived}''+``\emph{stuck}'', ``\emph{stuck}''+``\emph{stuck}'' and ``\emph{access}''+``\emph{access}'' gives rise to the best overall performance.  The second-best is [0110] ``\emph{arrived}''+``\emph{stuck}'' and ``\emph{stuck}''+``\emph{stuck}''.  Next is [1110] ``\emph{random movements}'', ``\emph{arrived}''+``\emph{stuck}'', and ``\emph{stuck}''+``\emph{stuck}'', and fourth is [0010] ``\emph{stuck}''+``\emph{stuck}''.  The following three combinations [1010], [0011] and [1011] also have relatively high `goodness', and confirm that the top eight all have [0010] `\emph{stuck}''+``\emph{stuck}'' enabled, and performance drops significantly without it.

The highest number of DNFs was 854/1000 (for [0001]), the lowest was 284/1000 (for [0111]), and there were 769/1000 DNFs for no cooperation at all ([0000]).  These results imply that up to 28\% of barrier configurations were unsolvable and $\sim$23\% were solvable \emph{without} cooperation, which means that $\sim$49\% were able to be solved \emph{with} cooperation.  The fact that [0001] resulted in a higher number of DNFs than [0000] implies that enabling the ``\emph{access}''+``\emph{access}'' strategy was actually detrimental to performance.

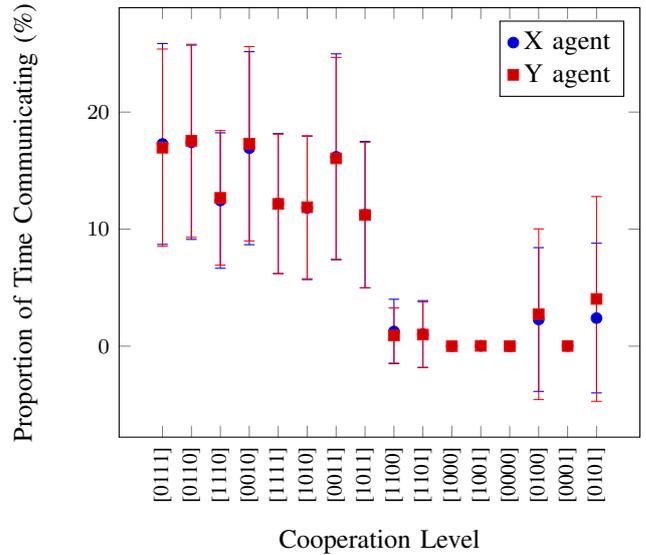
\begin{figure}[h!]
	\centering
	\begin{tikzpicture}
		\begin{axis}[x label style={at={(axis description cs:0.5,-0.1)},anchor=north},xlabel=Cooperation Level,ylabel=Proportion of Time Communicating (\%),xtick={0, 1, 2, 3, 4, 5, 6, 7, 8, 9, 10, 11, 12, 13, 14, 15},xticklabels={[0111], [0110], [1110], [0010], [1111], [1010], [0011], [1011], [1100], [1101], [1000], [1001], [0000], [0100], [0001], [0101]},xticklabel style={rotate=90}, legend pos=north east]
			\addplot+[only marks, error bars/.cd,
			y dir=both,y explicit]
			coordinates {
			(12, 0.0) +- (0.0, 0.0)
			(10, 0.0) +- (0.0, 0.0)
			(13, 2.27594) +- (6.14409, 6.14409)
			(8, 1.25356) +- (2.7534, 2.7534)
			(3, 16.9165) +- (8.24555, 8.24555)
			(5, 11.8167) +- (6.12704, 6.12704)
			(1,  17.4217) +- (8.29956, 8.29956)
			(2,  12.4432) +- (5.78908, 5.78908)
			(14,  0.0146092) +- (0.0308235, 0.0308235)
			(11,  0.0228188) +- (0.0247545, 0.0247545)
			(15,  2.39835) +- (6.39766, 6.39766)
			(9,  1.0331) +- (2.86055, 2.86055)
			(6,  16.179) +- (8.79654, 8.79654)
			(7,  11.2363) +- (6.25733, 6.25733)
			(0,  17.2781) +- (8.58085, 8.58085)
			(4,  12.1857) +- (5.98063, 5.98063)
			};
			\addplot+[only marks, error bars/.cd,
			y dir=both,y explicit]
			coordinates {
			(12, 0.0) +- (0.0, 0.0)
			(10, 0.0) +- (0.0, 0.0)
			(13, 2.73226) +- (7.29137, 7.29137)
			(8, 0.910321) +- (2.35392, 2.35392)
			(3, 17.2964) +- (8.30454, 8.30454)
			(5, 11.8721) +- (6.10759, 6.10759)
			(1,  17.5446) +- (8.24792, 8.24792)
			(2,  12.6717) +- (5.75424, 5.75424)
			(14,  0.0124633) +- (0.0392575, 0.0392575)
			(11,  0.0236578) +- (0.0259562, 0.0259562)
			(15,  4.03299) +- (8.75191, 8.75191)
			(9,  0.98792) +- (2.80034, 2.80034)
			(6,  16.0454) +- (8.62732, 8.62732)
			(7,  11.2077) +- (6.21543, 6.21543)
			(0,  16.9581) +- (8.4287, 8.4287)
			(4,  12.1537) +- (5.95948, 5.95948)
			};
			\legend{X agent,Y agent}
		\end{axis}
	\end{tikzpicture}
	\caption{Relationship between different combinations of cooperation and the \% of time the X and Y agents were communicating for matched agents in an environment containing three randomly placed barriers with random lengths and orientations.  The results are ordered by `goodness' from left to right.}
	\label{fig:comms}
\end{figure}

With regard to the proportion of time agents spent communicating, the results shown in \figurename~\ref{fig:comms} reveal that there is a clear relationship between the goodness of the cooperation combinations and the proportion of time the agents spent communicating.  As noted above, this is a function of whether [0010] ``\emph{stuck}''+``\emph{stuck}'' is enabled or disabled, and it clearly reflects the frequency with which situations containing local minima arise given the three random barriers.  It is also interesting to note that the highest levels of communication occurred for the top two cooperation combinations.

Finally, the correlation between the mean \% time spent communicating and mean solution times was consistently $\sim$0.4 for the best eight cooperation combinations.  This shows that harder barrier configurations required proportionally more inter-agent communications, as well as taking longer to solve.

\subsection{Mismatched Agents}

\begin{figure}[h!]
	\centering
	\begin{tikzpicture}
		\begin{axis}[
			xlabel=X Cooperation Level,
			ylabel=Y Cooperation Level,
    			colormap/hot,
			view={0}{90},
			xtick={0, 1, 2, 3, 4, 5, 6, 7, 8, 9, 10, 11, 12, 13, 14, 15},xticklabels={[0000], [1000], [0100], [1100], [0010], [1010], [0110], [1110], [0001], [1001], [0101], [1101], [0011], [1011], [0111], [1111]},
			xticklabel style={rotate=90},
			ytick={0, 1, 2, 3, 4, 5, 6, 7, 8, 9, 10, 11, 12, 13, 14, 15},yticklabels={[0000], [1000], [0100], [1100], [0010], [1010], [0110], [1110], [0001], [1001], [0101], [1101], [0011], [1011], [0111], [1111]},
			]
			\addplot3 [surf,shader=interp] table {
0	0	5.887962486
1	0	5.695290063
2	0	5.823678901
3	0	5.511129844
4	0	5.612941377
5	0	5.522842302
6	0	5.518770076
7	0	5.507709946
8	0	6.002405465
9	0	5.655295524
10	0	5.953840754
11	0	5.601136994
12	0	5.426198964
13	0	5.466234768
14	0	5.654508831
15	0	5.59994412
		
0	1	5.695290063
1	1	5.651531092
2	1	5.730440674
3	1	5.676719067
4	1	5.23097378
5	1	5.385242932
6	1	5.320253011
7	1	5.364181289
8	1	5.71275442
9	1	5.666356004
10	1	5.791671295
11	1	5.506719896
12	1	5.445904963
13	1	5.42783904
14	1	5.385050476
15	1	5.340575096
		
0	2	5.823678901
1	2	5.730440674
2	2	5.998772745
3	2	5.717893309
4	2	5.437866832
5	2	5.629428427
6	2	5.40276307
7	2	5.479570835
8	2	5.985472759
9	2	5.572336051
10	2	6.058524668
11	2	5.579498589
12	2	5.477032039
13	2	5.423796
14	2	5.452366842
15	2	5.371917053
		
0	3	5.511129844
1	3	5.676719067
2	3	5.717893309
3	3	5.295117858
4	3	5.022176264
5	3	5.274921638
6	3	4.937146068
7	3	5.146263762
8	3	5.698706748
9	3	5.533377905
10	3	5.708304898
11	3	5.334423274
12	3	5.223575494
13	3	5.2557929
14	3	5.180640918
15	3	5.100388438
		
0	4	5.612941377
1	4	5.676719067
2	4	5.437866832
3	4	5.022176264
4	4	5.062639607
5	4	4.876195844
6	4	4.800716333
7	4	4.878258804
8	4	5.334441412
9	4	5.177230252
10	4	5.510695506
11	4	5.149330512
12	4	4.889316144
13	4	5.039276113
14	4	4.889514806
15	4	4.675229039
		
0	5	5.522842302
1	5	5.385242932
2	5	5.629428427
3	5	5.274921638
4	5	4.876195844
5	5	4.901128799
6	5	5.016641044
7	5	4.918260634
8	5	5.539961796
9	5	5.390676725
10	5	5.416327989
11	5	5.082951715
12	5	5.019764686
13	5	5.045077015
14	5	5.015524715
15	5	4.92161005
		
0	6	5.518770076
1	6	5.320253011
2	6	5.40276307
3	6	4.937146068
4	6	4.800716333
5	6	5.016641044
6	6	4.878804982
7	6	4.809119428
8	6	5.44186458
9	6	5.472464864
10	6	5.350173547
11	6	4.895082476
12	6	4.790906053
13	6	4.948420223
14	6	4.766908334
15	6	4.8317497

0	7	5.507709946
1	7	5.364181289
2	7	5.479570835
3	7	5.146263762
4	7	4.878258804
5	7	4.918260634
6	7	4.809119428
7	7	4.978020792
8	7	5.499361701
9	7	5.349452063
10	7	5.607538488
11	7	5.03425766
12	7	4.900579545
13	7	5.072523133
14	7	5.011580388
15	7	4.859750913
		
0	8	6.002405465
1	8	5.71275442
2	8	5.985472759
3	8	5.698706748
4	8	5.334441412
5	8	5.539961796
6	8	5.44186458
7	8	5.499361701
8	8	5.980301889
9	8	5.481216579
10	8	5.928878983
11	8	5.59999525
12	8	5.424271419
13	8	5.478501225
14	8	5.569256194
15	8	5.488657488
		
0	9	5.655295524
1	9	5.666356004
2	9	5.572336051
3	9	5.533377905
4	9	5.177230252
5	9	5.390676725
6	9	5.472464864
7	9	5.349452063
8	9	5.481216579
9	9	5.81727091
10	9	5.653422047
11	9	5.862189866
12	9	5.292499243
13	9	5.496263347
14	9	5.384398983
15	9	5.41076013
		
0	10	5.953840754
1	10	5.791671295
2	10	6.058524668
3	10	5.708304898
4	10	5.510695506
5	10	5.416327989
6	10	5.350173547
7	10	5.607538488
8	10	5.928878983
9	10	5.653422047
10	10	5.914035297
11	10	5.627833139
12	10	5.589745858
13	10	5.408258087
14	10	5.558519045
15	10	5.653764917
		
0	11	5.601136994
1	11	5.506719896
2	11	5.579498589
3	11	5.334423274
4	11	5.149330512
5	11	5.082951715
6	11	4.895082476
7	11	5.03425766
8	11	5.59999525
9	11	5.862189866
10	11	5.627833139
11	11	5.535047554
12	11	5.16426124
13	11	5.305479948
14	11	5.130766376
15	11	5.209371991
		
0	12	5.426198964
1	12	5.445904963
2	12	5.477032039
3	12	5.223575494
4	12	4.889316144
5	12	5.019764686
6	12	4.790906053
7	12	4.900579545
8	12	5.424271419
9	12	5.292499243
10	12	5.589745858
11	12	5.16426124
12	12	4.964328309
13	12	5.035769341
14	12	4.873410055
15	12	4.901877318
		
0	13	5.466234768
1	13	5.42783904
2	13	5.423796
3	13	5.2557929
4	13	5.039276113
5	13	5.045077015
6	13	4.948420223
7	13	5.072523133
8	13	5.478501225
9	13	5.496263347
10	13	5.408258087
11	13	5.305479948
12	13	5.035769341
13	13	5.050266327
14	13	5.078722845
15	13	4.889560028
		
0	14	5.654508831
1	14	5.385050476
2	14	5.452366842
3	14	5.180640918
4	14	4.889514806
5	14	5.015524715
6	14	4.766908334
7	14	5.011580388
8	14	5.569256194
9	14	5.384398983
10	14	5.558519045
11	14	5.130766376
12	14	4.873410055
13	14	5.078722845
14	14	4.945472784
15	14	4.928725044
		
0	15	5.59994412
1	15	5.340575096
2	15	5.371917053
3	15	5.100388438
4	15	4.675229039
5	15	4.92161005
6	15	4.8317497
7	15	4.859750913
8	15	5.488657488
9	15	5.41076013
10	15	5.653764917
11	15	5.209371991
12	15	4.901877318
13	15	4.889560028
14	15	4.928725044
15	15	4.949594764
			};
		\end{axis}
	\end{tikzpicture}
	\caption{Heat map of the goodness measure for all combinations of X and Y cooperation for mismatched agents in an environment containing three randomly placed barriers with random lengths and orientations.  Blue corresponds to the best, and red to the worst.}
	\label{fig:MisM}
\end{figure}
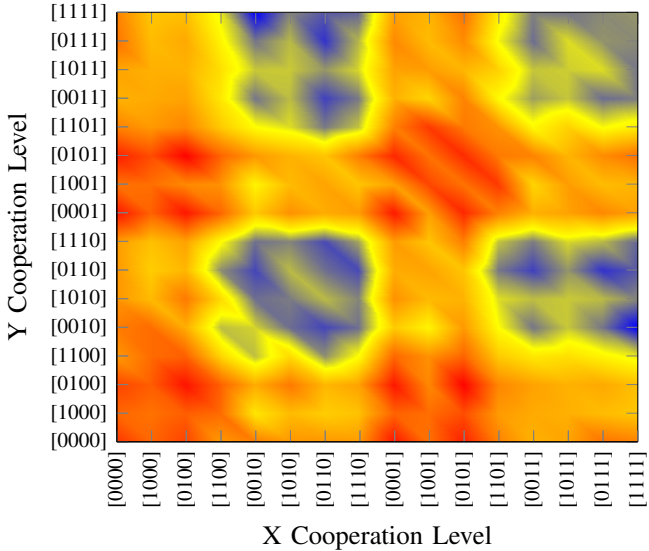

As an example of the consequences of allowing the cooperation level to be set independently for each agent, \figurename~\ref{fig:MisM} shows results for all combinations of matched and \emph{mismatched} agents in an environment containing three randomly placed barriers with random lengths and orientations.

The results of this experiment showed that up to 26\% of barrier configurations were unsolvable and $\sim$30\% were solvable \emph{without} cooperation.  This meant that $\sim$44\% were able to be solved \emph{with} cooperation.  

What is particularly interesting in the results shown in \figurename~\ref{fig:MisM} is that the good solutions are not confined to the diagonal, i.e.\ not restricted to the matched agents conditions.  In fact the outcomes resulting from the best mismatched agents are comparable to those for the best matched agents.  For example, the best performance over all cooperation combinations was obtained for [1111]+[0010] (i.e.\ where one agent had all cooperation modes enabled, and the other agent was only responding to ``\emph{stuck}''), and this result was slightly better than the best matched agents at [0110]+[0110].

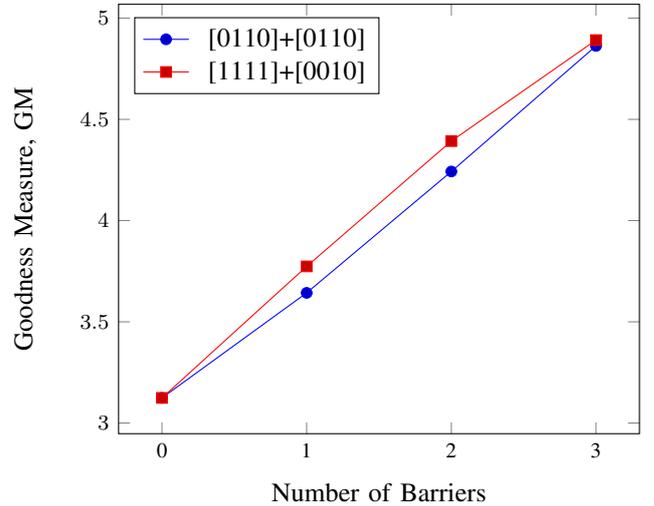
\begin{figure}[h!]
	\centering
	\begin{tikzpicture}
		\begin{axis}[xlabel=Number of Barriers,ylabel={Goodness Measure, GM}, xtick={0, 1, 2, 3}, xticklabels={0, 1, 2, 3}, legend pos=north west]
			\addplot+
			coordinates {
			(0, 3.12416)
			(1, 3.64301)
			(2, 4.24256)
			(3, 4.86268)
			};
			\addplot+
			coordinates {
			(0, 3.12417)
			(1, 3.77376)
			(2, 4.39272)
			(3, 4.89171)
			};
			\legend{\phantom{x}[0110]+[0110], \phantom{x}[1111]+[0010]}
		\end{axis}
	\end{tikzpicture}
	\caption{Relationship between the goodness measure and the number of random barriers for matched and mismatched agents.}
	\label{fig:GM0123}
\end{figure}

\begin{figure}[h!]
	\centering
	\begin{tikzpicture}
		\begin{axis}[xlabel=Number of Barriers,ylabel={Proportion of Time Communicating (\%)}, xtick={0, 1, 2, 3}, xticklabels={0, 1, 2, 3}, legend pos=north west]
			\addplot+
			coordinates {
			(0, 0)
			(1, 4.32329/2+4.29881/2)
			(2, 11.5836/2+11.4675/2)
			(3, 17.5992/2+17.5768/2)
			};
			\addplot+
			coordinates {
			(0, 0)
			(1, 3.61589/2+4.23405/2)
			(2, 8.82898/2+10.9701/2)
			(3, 12.6775/2+16.9292/2)
			};
			\legend{\phantom{x}[0110]+[0110], \phantom{x}[1111]+[0010]}
		\end{axis}
	\end{tikzpicture}
	\caption{Relationship between the number of random barriers and the total \% of time the X and Y agents were communicating for matched and mismatched agents.}
	\label{fig:COM0123}
\end{figure}
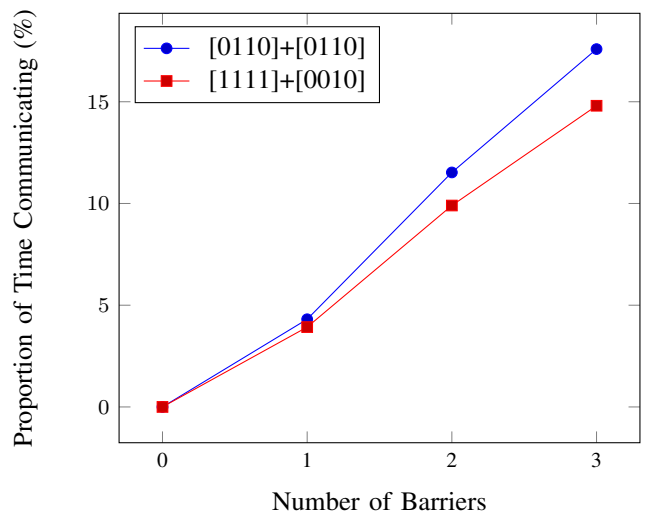

Further investigation into the consequences of allowing the cooperation level to be set independently for each agent was made by comparing the performance of the best matched and mismatched combinations mentioned above (i.e.\ [1111]+[0010] versus [0110]+[0110]) with varying numbers of barriers.  The results (shown in \figurename~\ref{fig:GM0123}) reveal that the matched agents performed slightly better than the mismatched agents.  However, as can be seen in \figurename~\ref{fig:COM0123}, the mismatched agents communicated less than the matched agents, with the difference being proportionally larger for the more difficult barrier configurations.

\section{Discussion}

Clearly, the task posed here is related to finding an optimal route on a map.  As such, two scenarios are possible: (i) the distance between the vehicle's current position and the target is known, but gradient descent may lead to a local minimum, or (ii) the distance between the vehicle's current position and the target is unknown due to an obstruction.  The first of these may be solved by \emph{planning} (assuming that the map is known), or by recognising that a local minimum has occurred and trying to jump out stochastically.  In the second scenario, only random search is possible.

However, this map-based analysis is based on the privileged perspective of a 2D agent.  In the task posed in this paper, the agents were purposefully designed to be 1D, precisely so that they did \emph{not} have access to a 2D map.  This meant that planning was not possible, and the recognition of arriving at a local minimum (or of simply not being able to see the target) required message-passing between the agents, i.e.\ explicit cooperation by communication.

Another insight to emerge from this work is the realisation that communication may be achieved by signalling (i.e.\ a `push' from a sending agent) or by observation (i.e.\ a `pull' from a receiving agent).  Clearly, the latter is less efficient due to the need for continuous monitoring.  Hence it can be said that, while an attention mechanism may be important, raising alerts in a timely manner are critical to success in a cooperative task.

It is also interesting to note that the overall paradigm is not specifically concerned with explicit message passing.  Given that the \emph{meanings} of the particular messages have implications for the subsequent behaviour, the scenario may also be viewed as one agent needing to appreciate the other's situation.  In other words, timely communications to overcome local minima in a cooperative problem space may be viewed as instantiating a primitive `theory-of-mind'\cite{Premack}.

\section{Summary and Conclusion}

This paper has addressed the question as to what conditions the timing and structure of communication in continuous cooperative interaction.  Experiments have been conducted using a PCT-based simulation of a cooperative task in which two independent one-dimensional agents are obliged to communicate in order to solve a two-dimensional path-finding problem.  

Results from a number of simulation experiments have confirmed the hypothesis that appropriately timed communication between agents can overcome local minima in a joint problem space.  It has also been shown that asymmetric levels of cooperative communication can be as effective as equally matched partners, and can even reduce the level of communications required to achieve the same level of performance.

Finally, although this study was aimed at \emph{extrinsic} communication between multiple agents, it is interesting to note that the results also apply to \emph{intrinsic} communications within a single agent.

\newpage

\bibliographystyle{IEEEtran}
\bibliography{IEEEabrv,RKM-AISB23}

\end{document}